\documentclass{svproc}
\usepackage{lipsum}

\usepackage[left=1in,right=1in,top=1.5in,bottom=1.5in]{geometry}
\usepackage{url}
\usepackage[ruled,vlined]{algorithm2e}
\usepackage{lipsum}
\usepackage{afterpage}
\usepackage{enumitem}
\usepackage{caption}
\usepackage{mathtools}
\usepackage{subfig}

\begin{document}

\mainmatter              
\title{A Vision Based Deep Reinforcement Learning Algorithm for UAV Obstacle Avoidance}

\titlerunning{UAV Obstacle Avoidance}  
%
\author{Jeremy Roghair \and Kyungtae Ko \and
Amir Ehsan Niaraki Asli \and Ali Jannesari*}

\authorrunning{}
\authorrunning{Roghair et al.} 
\institute{Department of Computer Science, Iowa State University, IA, USA,\\
\email{\{jroghair, kyungtae, niaraki, jannesari\}@iastate.edu}}

\maketitle           

\begin{abstract}
Integration of reinforcement learning with unmanned aerial vehicles (UAVs) to achieve autonomous flight has been an active research area in recent years. An important part focuses on obstacle detection and avoidance for UAVs navigating through an  environment. Exploration in an unseen environment can be tackled with Deep Q-Network (DQN). However, value exploration with uniform sampling of actions may lead to redundant states, where often the environments inherently bear sparse rewards.  To resolve this, we present two techniques for improving exploration for UAV obstacle avoidance. The first is a convergence-based approach that uses convergence error to iterate through unexplored actions and temporal threshold to balance exploration and exploitation. The second is a guidance-based approach using a Domain Network which uses a Gaussian mixture distribution to compare previously seen states to a predicted next state in order to select the next action. Performance and evaluation of these approaches were implemented in multiple 3-D simulation environments, with variation in complexity. The proposed approach demonstrates a two-fold improvement in average rewards compared to state of the art.
\keywords{UAV, Reinforcement Learning, Object Detection, Exploration, Obstacle Avoidance}
\end{abstract}
\section{Introduction}
%
Autonomous flight for UAVs (unmanned aerial vehicles) has become an increasingly important research area in recent years. One of the major challenges with autonomous motion planning is ensuring that an agent can efficiently explore a space while avoiding collision with objects in complex and dynamic environments. To resolve this, recent research has turned to applying deep reinforcement learning techniques to robotics and UAVs 
\cite{xie2017towards}, \cite{inproceedings}.

Reinforcement learning is a way for an agent to navigate through an environment, interact with that environment and use the observed rewards gained for each action it takes as feedback to optimize a policy. This iterative feedback loop of receiving signals through a sensor to reflect the agents current state allows it to train an optimal policy. Part of the research in this area has focused on employing a single sensor such as a camera \cite{xie2017towards}, \cite{10.1145/1102351.1102426} to control an agent, while others have focused on combining the signals from multiple sensors \cite{Preiss2017CrazyswarmAL}. Object detection algorithms have shown tremendous success in the past few years for both detection and segmentation in computer vision. These methods have recently been incorporated with reinforcement learning and applied to UAVs for autonomous path planning \cite{asli2019energy}, landing \cite{inproceedings}, and obstacle avoidance \cite{Ma2018ASR}, \cite{Smolyanskiy2017TowardLA}. 

The design of a robust learning algorithm to develop the agent's behavioral policy, is challenging. Particularly, when an agent navigates through an environment and learns from the state transitions, it may rarely receive any significant reward or signal from the environment in which to learn from, resulting in a significantly slower learning rate\cite{Gou2019DQNWM}. The Dueling Double Deep Q Network (D3QN) algorithm for obstacle avoidance introduced by \cite{xie2017towards} uses the $\epsilon$-greedy or Deep Q-Network (DQN) approach introduced by Mnih et al. \cite{Mnih2013PlayingAW} for choosing between exploration or exploitation. That is, either exploring through a stochastic process (exploration) or exploiting the behavioral policy learned so far in order to chose the next action. In the case of $\epsilon$-greedy exploration, choosing the next action is done through uniform random sampling, however, this choice may inevitably lead to states that are too similar or redundant to those recently visited and eventually lead to slow learning and significantly inflated training time. 

In this work, we focus on solving these challenges with regards to obstacle avoidance for UAVs through improved exploration using only a stereo camera within a simulation environment. The main contributions of this paper are as follows:

\begin{itemize} 
\itemsep0em 
\item Extended D3QN obstacle avoidance algorithm to UAVs from 2-dimension to 3-dimension environments.
\item Integrated object detection towards improving depth estimation for obstacle avoidance of static and non-static objects.
\item Configured and visualized 3 dimensional simulation environments for deep reinforcement learning.
\item Introduced the domain network and Gaussian mixture model for guidance exploration which exhibited more efficient learning for complex environments compared to conventional D3QN and convergence exploration.
\end{itemize}
\section{Related Work}
\label{sec:related-work}

Here, a basic underlying background on reinforcement learning with a Q-network is provided along with milestone studies on DQN for robot exploration with remarks on obstacle avoidance.   

\subsection{Reinforcement Learning for Obstacle Avoidance}

Deep Reinforcement learning applied to a UAV agent is a process of training to navigate the UAV through an environment with actions and corresponding rewards. An agent interacts with an environment through a Markov decision process from its current state $s_{t}$ by performing an action $a_{t}$ to transition to a new state $s_{t+1}$ in order to receive reward $r_{t}$. Generally, these states are represented as depth prediction images and the size of the reward influences the strength of the signal for which the UAV will alter its behavior policy. For the problem of obstacle avoidance, \cite{xie2017towards} demonstrated that deep Q-learning as introduced by \cite{wang2015dueling}, is effective for robotic navigation in 2-Dimensions. Alternative approaches have been tried including work by Khan et al. \cite{Kahn2017UncertaintyAwareRL} provided a model-based approach that estimates the probability that the UAV would have a collision within an unknown environment, allowing for specific actions depending on the certainty of the prediction. However, the generalization of this approach is questionable as the tested scenarios were applied primarily to static obstacles. Additionally, Long et al. \cite{Long2017TowardsOD} worked towards the problem of collision of non-stationary objects by developing a decentralized framework for multiple robots navigating through an environment, although this approach wasn't explicitly tested for UAVs. The usefulness of deep reinforcement learning, specifically DQN for UAVs was articulated by \cite{Wang2017AutonomousNO}, demonstrating the usefulness of combining it with a actor-critic paradigm, that is one that uses two deep learning networks, one for short and long term decision making in order to optimally choose actions as it traverses the environment. The D3QN algorithm introduced by \cite{xie2017towards} utilized DQN with the dual network approach for obstacle avoidance, which evaluated a state-action pair by its estimated Q-value (Q($s_{t}$,$a_{t}$)) reflecting its accumulated reward so far, the instantaneous reward of transitioning to the next state $s_{t+1}$ and its expected future rewards from this new state. Equation \ref{ref:equation1} shows how state-action values are updated for a Q-network. Where, $\alpha$ is the learning rate and $\gamma$ is discount factor~\cite{Sutton1998}.

\begin{equation} 
\label{ref:equation1}
Q'(s_t,a_t) = Q(s_t,a_t) + \alpha (r_{t}+\gamma \cdot max_{a}Q(s_{t+1},a) - Q(s_t,a_t))
\end{equation}

\subsection{Exploration}
Making improvements to exploration for deep reinforcement learning algorithms that use the actor-critic paradigm \cite{10.5555/3016100.3016191} has been actively researched due to the tumultuous nature of environments having sparse rewards such as those for UAVs. Since the inception of the work by Mnih et al. on DQNs \cite{Mnih2013PlayingAW} various approaches have been explored to make improvements over the uniform random selection of actions ($\epsilon$-greedy) where the agent takes a random exploratory action with the probability of $\epsilon$ and exploits the policy otherwise. One of the key insights of improving exploration of an agent focuses on making improvements to the way transitions are sampled from the replay memory. With this insight Schaul et al. \cite{Schaul2015PrioritizedER} showed that partitioning the replay memory according to temporal difference error can assist in the generation of new distributions for state transitions to sample from. This results in a more efficient learning and a faster convergence. Therefore, using a prioritized replay can lead to states that are sparse and seen more infrequent. Leading to states that are rarer allows the agent a greater opportunity to learn than previously seen states. Oh et al. \cite{Oh2015ActionConditionalVP} demonstrated this in his work, by producing a prediction of the next 100 possible states and choosing an action based on comparison to previously seen states using a Gaussian kernel. Similarly, Gou et al. \cite{Gou2019DQNWM} demonstrated a similar approach that uses an dynamic network to make a prediction for the next state using a multivariate Gaussian. Each demonstrated that a prediction of future states combined with a similarity metric was effective at improving learning efficiency in environments with sparse rewards. 


\section{Methodology: Towards Improving Exploration}

In this section, we introduce two different exploration approaches with the goal of allowing a UAV to learn an environments dynamics quicker than $\epsilon$-greedy exploration resulting in more efficient obstacle avoidance. 

\subsection{Training Setup}
\label{sec:trainingsetup}

The purpose of our experiments is to train a UAV to move autonomously through an environment, while maximizing its flight duration and coverage area of the state space in order to sufficiently learn how to avoid collisions. To this end, we use two simulation environments for our experiments, one simple and one complex. 

\begin{figure}
\centering
    \includegraphics[width=12cm]{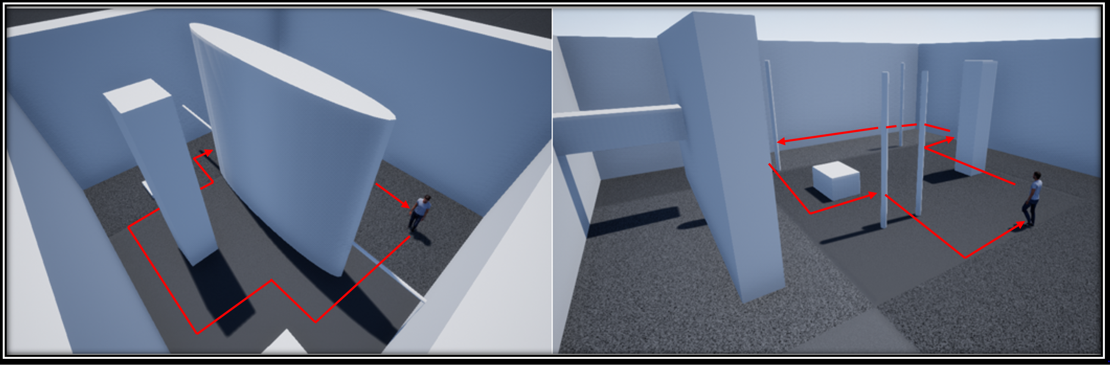}
  \caption{Simulation Environments: Simple (left) \& Complex (Right)}
  \label{fig: environments}
\end{figure}

\begin{figure}
\centering
    \includegraphics[width=8cm]{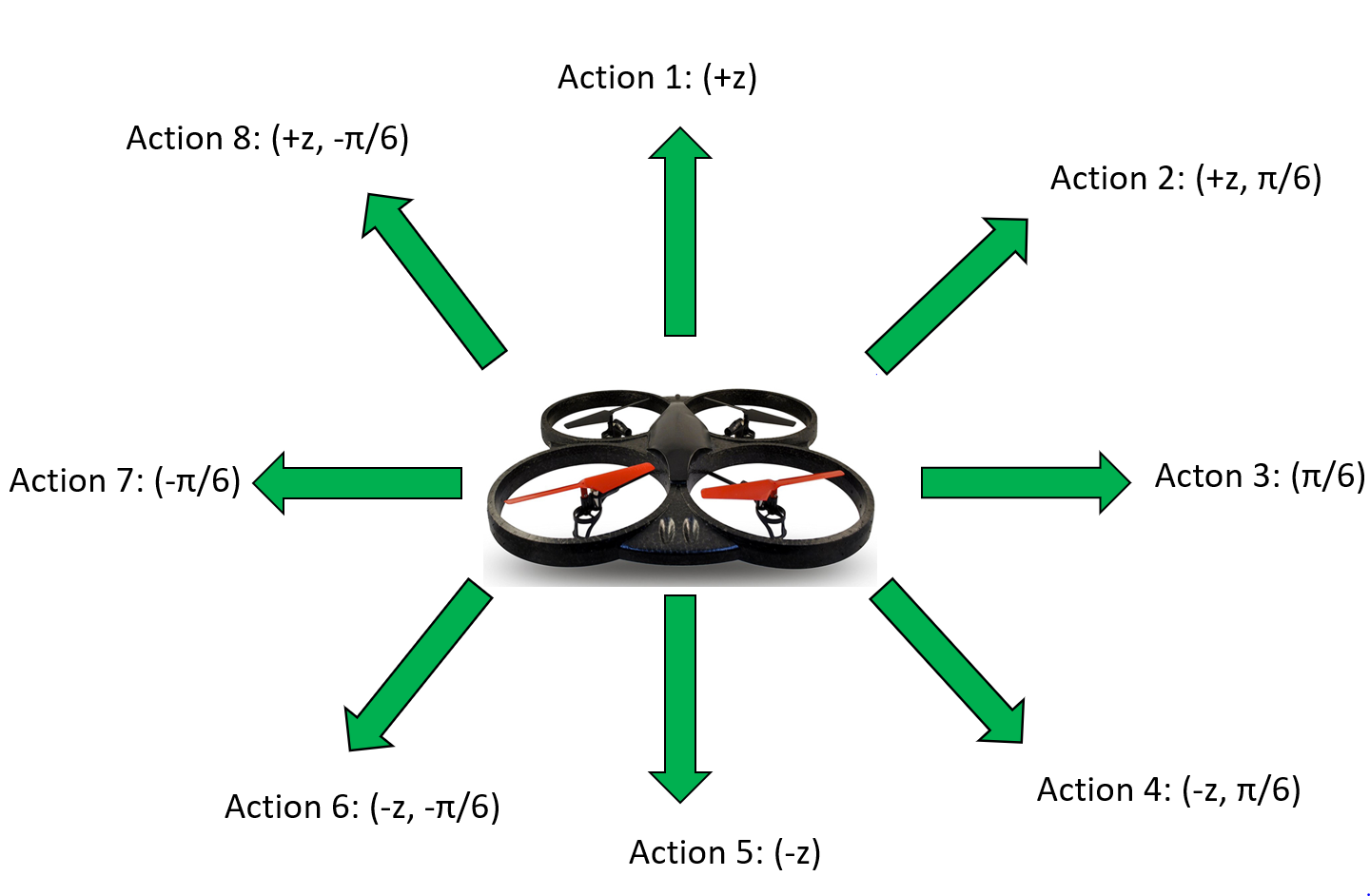}
  \caption{UAV Angular Velocity Actions}
  \label{fig: uavmovement}
\end{figure}

A multi rotor drone model including its navigation, control and physics of its environment are implemented by using AirSim \cite{Shah2017AirSimHV}. Typical UAV movement that uses yaw, roll and pitch have convenient methods within AirSim for transformations from inputs such as linear and angular velocity. There are a total of ten actions, two actions affect the linear velocity ($v$) of the UAV and the remaining actions change its angular velocity ($\psi$) in 3 dimensions. In our experiments our UAV can move with linear velocity of $1.2$ m/s or $0.6$ m/s (z-directional) and it can turn in any forward facing direction with an angular velocity of $\pi/6$. Figure \ref{fig: uavmovement} illustrates the angular velocity actions that it can take from its current state $s_{t}$. 

Our deep learning model operates with two modules. A Policy module for our deep Q-network and object detection model, and an Interaction module for the communication between the agent and the simulation environment. The Policy module is composed of an online network and target network which follow an actor-critic paradigm as \cite{Sutton1998}. The interaction module receives camera input of an RGB image, processes it with object detection, and uses the resulting bounding box of a detected non-static objects to improve the depth estimation image used by the UAV to learn the environment. This improved depth image is processed through the Policy module to determine an estimated $Q(s_{t},a_{t})$ for each possible action $a_{t}$ in the current state $s_{t}$. The agent takes the action $a_{t}$ from its current state with the largest Q-value, i.e. $argmax_{a}Q(s_{t},a_{t})$ in order to transition to a new state $s_{t+1}$. These state-action transitions $(s_{t}, a_{t}, r_{t}, s_{t+1})$ are saved in a replay memory $M$, in order to optimize our target network which provides an estimated optimal Q-value that the UAV strives towards. This replay memory saves approximately  $5,000$ most recent transitions. 

\begin{figure}[h]
  \center
  \includegraphics[width=9cm,scale=0.7]{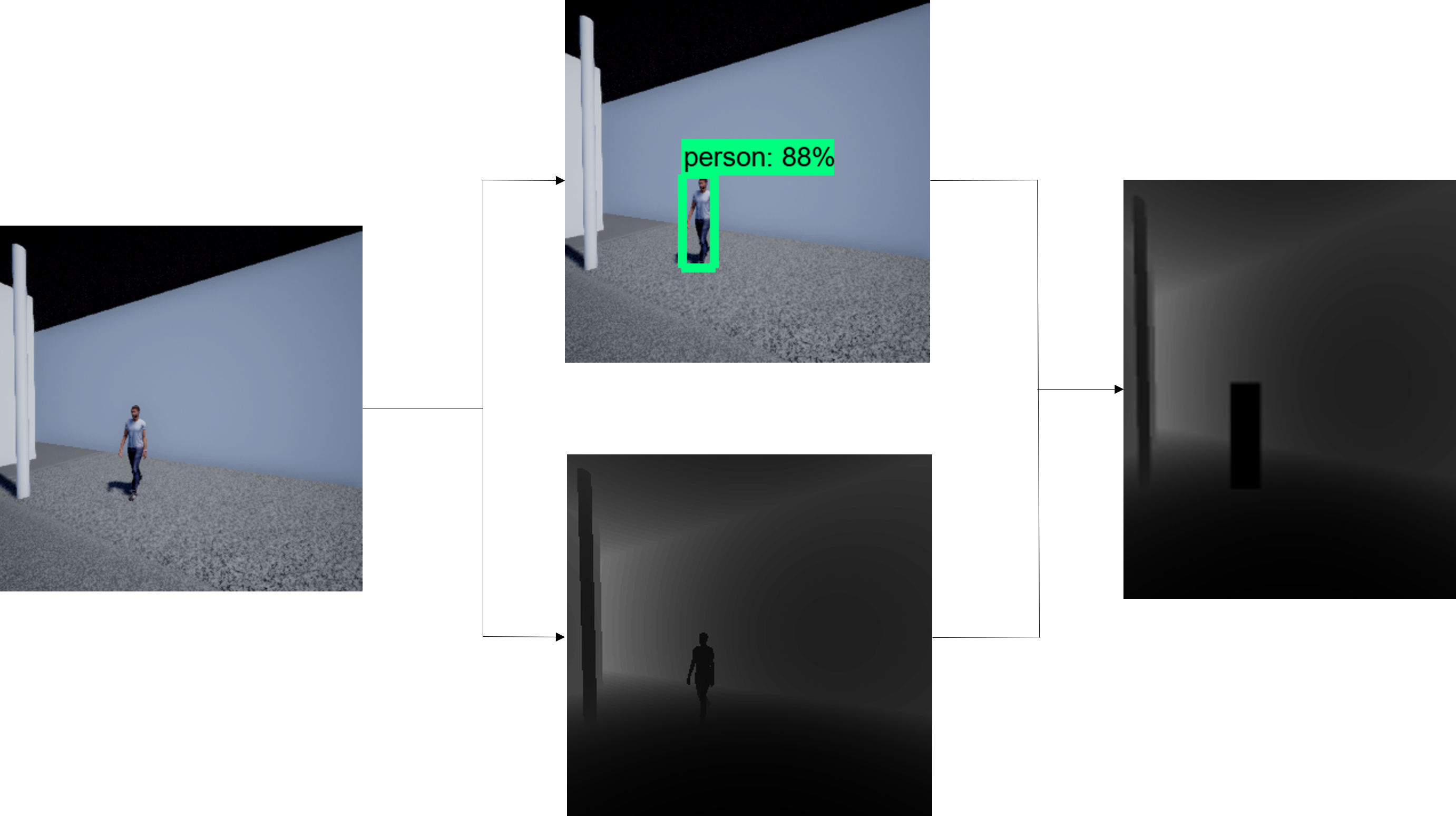}
  \caption{Improved Depth Estimation Using Object Detection}
  \label{fig: depthestimation}
\end{figure}
\begin{figure}[h]
  \centering
  \includegraphics[width=15cm]{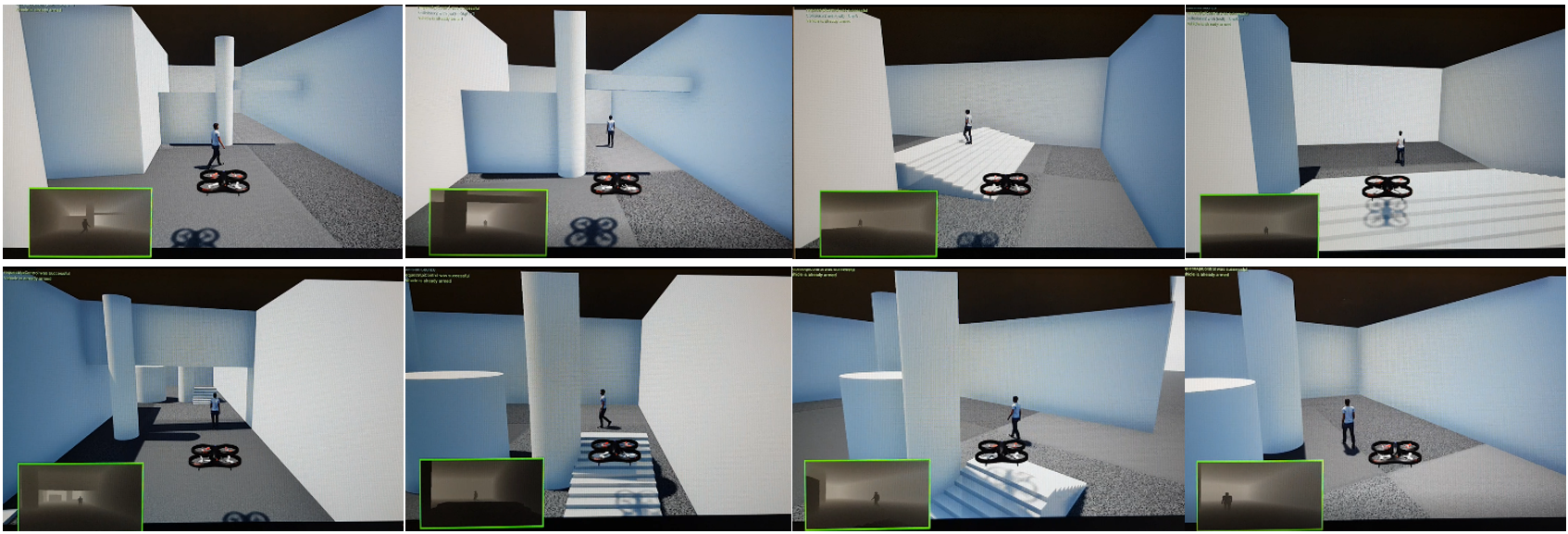}
  \caption{Training with Simulation Environment}
  \label{fig: trainingairsim}
\end{figure}
\vspace*{-\baselineskip}

Figure \ref{fig: depthestimation} shows how RGB images are pre-processed with bounding boxes of detected objects to modify the depth image received from our stereo camera in order to improve depth estimation of non-static objects (a person in our experiments). By providing the UAV with the illusion that the person is closer than it actually is, it incentives the UAV to learn to change its trajectory away from the person sooner to avoid collision. We assume in our experiments that the person is always ahead of the UAV. 
The goal of the UAV is to maximize expected future rewards and to do so, it tries to move through the environment as fast as possible while covering as much of the state space as it can without collision. Equation 2 gives the reward function for all the states, except if it is terminal. Each episode terminates, if the UAV collides with an object (reported by the interaction module) or if the agent takes a maximum of 500 steps.

\begin{equation} 
r_{t} = v \times \cos(\psi) \times \delta t +  (\lambda \times BB_{distance})-(\rho \times BB_{penalty})
\end{equation}
The UAV navigates through the environment as fast as it can by being rewarded based on its distance traveled $(v \times \cos(\psi))$. It avoids a person, by being rewarded if the person appears within its field of vision and is too further away from the center ($BB_{distance}$)in order to incentivise the UAV to turn away from the person. Moreover, the UAV is negatively rewarded the closer it comes to the person by measuring the aspect ratio of the bounding box ($BB_{penalty}$), the larger it is the closer the person is and the larger the negative reward will be. If the UAV does collide with an object it receives a reward of $-10$. The reward function is a heuristic that integrates object detection to improve avoiding obstacles. Therefore, it relies on hyper-parameters ($\delta t,\lambda , \rho$) that must be tuned during training to ensure rewards are adequate and learning is sufficient towards the goals of the UAV. Figure \ref{fig: trainingairsim} shows the UAV during training. 
\vspace*{-1cm}
\subsection{Convergence Exploration}
\vspace*{-1cm}
A fundamental issue with $\epsilon$-greedy exploration is that it relies on random chance to choose its next action, many times resulting in an agent falling into a local minimum of repeating the same action for short sighted short-term rewards. To address this, we explore convergence exploration introduced by Masadeh et al. \cite{masadeh18} for improving the task of obstacle avoidance.

\begin{minipage}[t]{0.8\linewidth}
   \begin{algorithm}[H]
    \caption{Convergence Exploration}
    \label{alg:convergence}
    \SetAlgoLined
  Initialize replay memory $M$ \;
  Initialize Online Network $Q$ with random weights $\theta$\;
  Initialize Target Network $Q_{tar}$ with random weights $\theta_{tar}$\;
  Initialize $\epsilon_{goal}$ \;
  \For{episode: 1 to E}{
  	\While{Episode is not reset}{
		Explore $=$ True if training step $< \tau$ \;
		\eIf{Explore}{
            Pick $a_{t}$ uniformly at random \;
            \While{($y_{t} - Q(s_{t}, a_{t}|\theta))^2 < \zeta$}{
                Pick $a_{t}$ uniformly at random\;
            }
		}{
			Pick $a_{t} = argmax_{a}Q(s_{t}, a|\theta)$\;
		}
		Execute $a_{t}$, observe reward $r_{t}$\;
		Transition to state $s_{t}$ using $a_{t}$ and observe reward $r_{t}$\;
		M = M $\cup (s_{t}, a_{t}, r_{t}, s_{t+1})$ \;
		Sample $b$ from $M$ \;
		\For{$i$ in $b$}{
		 \eIf{Terminal}{
		    $y_{i} = r_{i}$\;}
		   {$y_{i} = r_{i} + \gamma max_{a}Q_{tar}(s_{i},a|\theta_{tar})$}
		Optimize $Q$ w.r.t $\theta$ on ($y_{i} - Q(s_{i}, a_{i}|\theta))^2$\;   
		}
		Set $Q_{tar} = Q$\;
		$\epsilon$ = $\frac{\epsilon_{0}-\epsilon_{goal}}{episode}$\;
  	}
  }  
 \end{algorithm}
\end{minipage}%

In convergence exploration, a state-action familiarity or level of convergence is examined to decide between exploration and exploitation. Two parameter, $\tau$ and $\zeta$ act as a threshold between exploration time and minimum convergence error for exploitation. If the total training time is $T$, the agent is forced to explore during a duration of $\tau$, then exploit in remaining time $T- \tau$. During exploration, the parameter $\zeta$ checks if the convergence error of taking an action is less than $\zeta$, indicating the action for the current state sufficiently converged. If the algorithm judged that the action has converged, i.e. been sufficiently explored, then another random action will be explored until this new action has also converged. Therefore, the agent can deal with unexplored states and untried actions by iteratively searching over each state action pair resulting in faster learning. Masadeh et al. \cite{masadeh18} showed that convergence exploration outperforms $\epsilon$-greedy \cite{Mnih2013PlayingAW} in a discreet 2-Dimensional environment. The full algorithm can be seen in algorithm \ref{alg:convergence}.


\subsection{Guidance Exploration}

The problem of sparse rewards can be resolved in another way rather than convergence and $\epsilon$-greedy exploration. Instead, we can utilize a guidance based approach, inspired by \cite{Gou2019DQNWM}. We introduce a new method called domain network to perform a one step ahead prediction of the next state and compare the predicted state with the previously seen states in the replay memory. Subsequently, the chosen action will be the one that leads to the state with the least similarity to the previously experienced states. Our domain network utilizes a CNN architecture inspired by the encoding network introduced by Pathak et al. \cite{pathakICMl17curiosity} rather than the dynamic network utilized by Gou et al. \cite{Gou2019DQNWM} and uses a refined similarity measurement between the predicted state and previously visited states. For the similarity measurement, we applied a Bayesian Gaussian mixture in comparison to \cite{Gou2019DQNWM}, \cite{Oh2015ActionConditionalVP} which utilized a joint multivariate Gaussian and Gaussian kernel respectively. 

\begin{minipage}[t]{0.8\linewidth}
   \vspace{5pt}
    \begin{algorithm}[H]
    \caption{Guidance Exploration}
    \label{alg:guidance}
    \SetAlgoLined
  Initialize rank based prioritized replay memory $M$ \;
  Initialize Online Network $Q$ with random weights $\theta$\;
  Initialize Target Network $Q_{tar}$ with random weights $\theta_{tar}$\;
  Initialize Domain Network $D$ with random weights $\theta_{d}$\;
  Initialize $\epsilon_{goal}$ \;
  \For{episode: 1 to E}{
  	\While{Episode is not reset}{
		Explore $=$ True with probability $\epsilon$ \;
		\eIf{Explore}{
            Sample $V$ from $M$\;
            $S_{V} = S_{V} \cup V$\;
            Fit BGMM $\mathcal{G}$ to $S_{V}$ \;
            Execute $a_{t} = argmin_{a}\mathcal{G}(D(s_{t}, a|\theta_{d}))$\;
		}{
			Pick $a_{t} = argmax_{a}Q(s_{t}, a|\theta)$\; }
		Execute $a_{t}$, observe reward $r_{t}$\;
		Transition to state $s_{t}$ using $a_{t}$ and observe reward $r_{t}$\;
		M = M $\cup (s_{t}, a_{t}, r_{t}, s_{t+1})$ \;
		Sample $b$ from $M$ \;
		\For{$i$ in $b$}{
		 \eIf{Terminal}{
		    $y_{i} = r_{i}$\;}
		   {$y_{i} = r_{i} + \gamma max_{a}Q_{tar}(s_{i},a|\theta_{tar})$}
		Optimize $Q$ w.r.t $\theta$ on ($y_{i} - Q(s_{i}, a_{i}|\theta))^2$\;   
		Optimize $D$ w.r.t $\theta_{d}$ on $(S_{i+1}- D(S_{i},a_{i}|\theta_{d}))^{2}$\;
		Sort $M$ by TD error \;
		}
		Set $Q_{tar} = Q$\;
		$\epsilon$ = $\frac{\epsilon_{0}-\epsilon_{goal}}{episode}$
  	}
  }
  \end{algorithm}
\end{minipage}

The domain network functions to provide a prediction of the future next state $s_{t+1}$ from the current state $s_{t}$. The network's architecture uses a state-action encoding as input into a fully connected feed-forward network to make a prediction of the next state for each available action. The weights of the network are periodically updated from previous transitions, sampled from the replay memory. As suggested by \cite{Schaul2015PrioritizedER}, we implemented a rank based prioritized replay memory. 
Sampling from a rank based memory replay of previously visited states ensures that the sampled states are not as similar to one another, resulting in a wider variation of less redundant states to learn from.

After a prediction of the next state $s_{t+1}$ by the domain network, it is evaluated for similarity to a sample $V$ of previously visited states. We determine the similarity of the predicted state with the states in $V$ by deriving an estimated probability distribution $\mathcal{G}$ from $V$ in order to produce a probabilistic interpretation of the next action our UAV should take. Rather than modeling this distribution as a joint Gaussian with an empirical mean and co-variance for all states in $S_{V}$ such as in \cite{Gou2019DQNWM}, we model ours as a Bayesian Gaussian mixture (BGMM) in which each component $k$ of $S_{V}$ has its own mean and co-variance. The Gaussian mixture distribution $\mathcal{G}$ of $S_{V}$ is therefore defined as:

\begin{equation} 
\mathcal{G} = \sum_{k=1}^{m}\alpha_{k} \mathcal{N}(s|\mu_{k},\textstyle \sum_{k})
\end{equation}

 In our experiments, states are represented as pixels, hence a Gaussian distribution is fit to patches or sub-populations of these states. A Gaussian mixture is useful for modeling sub-populations of previously visited states in a dynamic environment rather than as a joint distribution across all states. Inspiration for using a Gaussian Mixture Model to model the distribution of past states was inspired by related work modeling images to aid in object tracking and tracking \cite{dadi13IJSCE}, \cite{Chavan2017MultipleOD}.
 

\section{Results and Discussion}

The results of the simulation tests for the new exploration algorithms were evaluated based on the goals of the UAV, i.e. cumulative reward and covered area of the environment. To this end, we evaluated the UAV by analyzing how well it performed towards these goals by looking at average rewards and training steps covered over a block of episodes. Here we define a block of episodes as 100 training episodes per block. Moreover, in order to examine how effective these new exploration algorithms were for obstacle avoidance during training, we compared the performance to D3QN introduced by Xie et al \cite{xie2017towards}.

\setlength{\textfloatsep}{5pt} 
\begin{figure}[h]%
  \centering
  \subfloat[Simple Environment]{\includegraphics[width=0.5\textwidth]{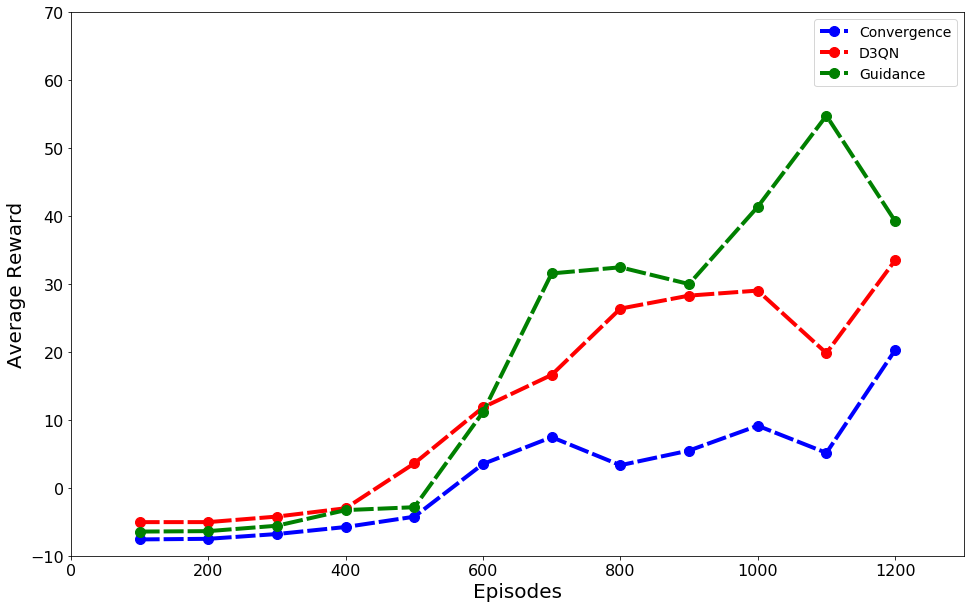}\label{fig: averagerewards1}}
  \hfill
  \subfloat[Complex Environment]{\includegraphics[width=0.5\textwidth]{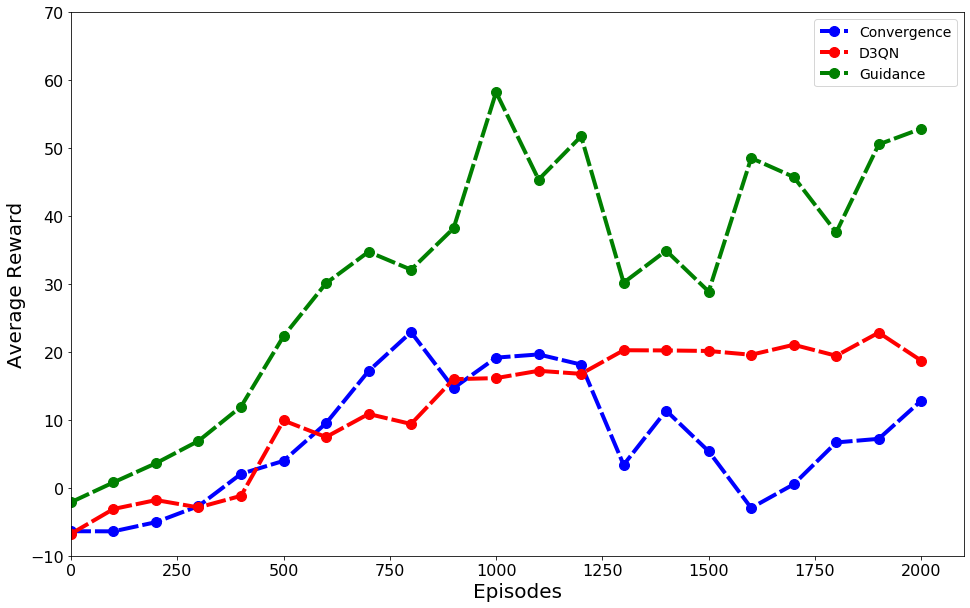}\label{fig: averagerewards2}}
  \caption{Accumulated Average Reward in Simulation Environments}
  \label{fig: averagerewards}
\end{figure}

As shown in figure \ref{fig: averagerewards}, the guidance exploration algorithm out performs D3QN and convergence exploration for each environment in terms of average reward. In each environment, convergence exploration fails to learn an optimal policy faster than D3QN, where as guidance exploration finds an optimal policy sooner. This is likely due to guidance exploration reaching new and unseen states sooner than D3QN and convergence exploration, resulting in learning the dynamics of the environment quicker. Related to this, there are fluctuations of the average reward over the course of the duration of a block of episodes many of which are likely due to the UAV reaching new sections of the environment it either hasn't seen before or frequently enough before to learn sufficiently how to navigate the space. This is especially evident in the complex environment in figure \ref{fig: averagerewards2} between episodes 1200 and 1500. After the environment is sufficiently explored, the UAV learns a more robust optimal policy and continues to accumulate reward.  

Moreover, the results of the average reward of the complex environment is more significant than the simple environment. In the complex environment you can clearly see from figure \ref{fig: averagerewards2} that early on that the guidance exploration algorithm is trending towards a more optimal policy sooner than the simple environment in figure \ref{fig: averagerewards1}. One possible explanation could be that the simple environment is smaller, with less objects to collide and interact with. Therefore it is a more predictable environment such that alternative exploration approaches aren't as beneficial compared to those with larger state spaces and even sparser rewards as is the case with the complex environment.  

Additionally, convergence exploration failed to learn both environments as shown in figure \ref{fig: averagerewards}. In 2-dimensional environments with discrete states, convergence-exploration was able to outperform DQN \cite{masadeh18}, however it fails to adapt to 3-dimensional environments with larger state spaces in comparison to duel deep q-networks such as D3QN and guidance exploration. Convergence exploration tends to be biased towards exploration in early stages of training until full convergence of each action is completed. Therefore, a low rate of exploitation of best actions tends to lead to poor performance during training. 

\begin{figure}[h]%
  \centering
  \subfloat[Simple Environment]{\includegraphics[width=0.5\textwidth]{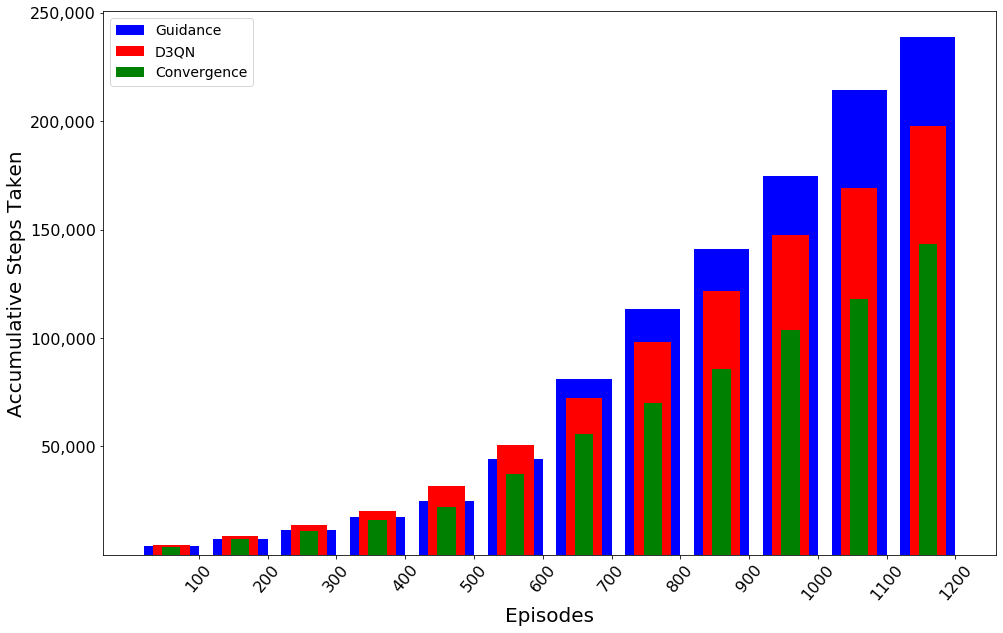}\label{fig: trainsteps1}}
  \hfill
  \subfloat[Complex Environment]{\includegraphics[width=0.5\textwidth]{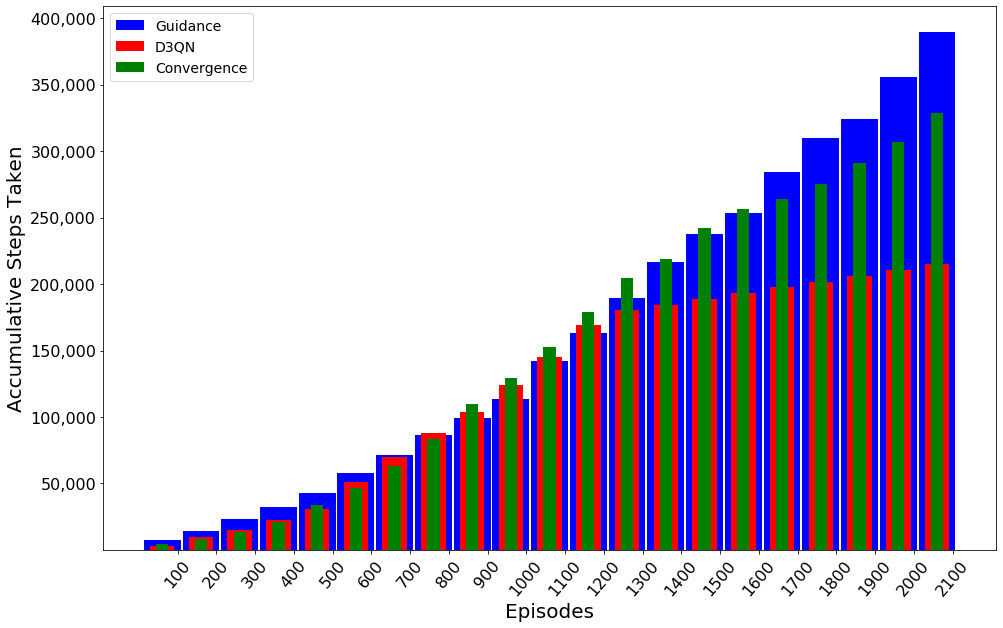}\label{fig: trainsteps2}}
  \caption{Steps taken prior to termination in simulation environments}
  \label{fig: trainsteps}
\end{figure}

The primary goal of the UAV is to explore as much of the state space as possible, for as long as possible. The UAV stops only if it collides with an object or if it reaches 500 steps and has converged during the duration of a training episode. Therefore, to evaluate how well our algorithm is tending towards this goal we evaluate the training steps taken over the course of episode blocks in figure \ref{fig: trainsteps}.

Each environment demonstrates again that guidance exploration outpaces D3QN and convergence exploration towards the goal of exploring more of the state space. For the complex environment, one can infer from figure \ref{fig: trainsteps2} that guidance exploration is trending towards less collisions and a significantly higher convergence rate in comparison to D3QN. Particularly for the last 100 episodes, the agent with guidance exploration policy takes and average of 350 steps per episode prior to termination when the number of steps taken for the agent with D3QN is an average of 50. This same effect is present in the simple environment but less pronounced. 

An interesting observation of convergence exploration is that its steps taken and hence states explored is significantly lower than D3QN in the simple environment (figure \ref{fig: trainsteps1}), but higher in the complex environment (figure \ref{fig: trainsteps2}). However, the average reward is lower in the complex environment for later episode blocks for convergence exploration as seen in figure \ref{fig: averagerewards2}. This could be due to the algorithm learning a policy which results in lower reward but takes significantly more actions that tend towards navigating as fast as possible through the environment. 

Overall, guidance exploration demonstrates nearly double the performance compared to D3QN and convergence exploration in the complex environment and roughly one and a half times the performance in the simple environment. As noted by Xie et al \cite{xie2017towards}, the benefit of D3QN was to aid in the overestimation of Q values inherent in the original DQN architecture and that improving exploration may not alleviate this issue. However, given the performance of these other exploration approaches, it appears that an agent making a more refined exploration choice early on does lead to faster convergence towards an optimal policy. 

The reward function presented here, is a heuristic based on goals that are defined for the UAV. The specific heuristic introduced in section 4 has hyper-parameters for the distance traveled ($ \delta$), bounding box distance ($\lambda$) and aspect ratio ($\rho$). These hyper-parameters reduce our ability to generalize our approach of incorporating object detection with deep reinforcement learning since applying our UAV towards other tasks could require a different reward function would need to be derived and tuned so that it tends towards those new objectives in an scalable way. 

The exploration approaches were compared to the D3QN obstacle avoidance algorithm since it is one of the most recent state-of-the-art algorithm for this task. Since our algorithms improve exploration that eventually propagates to more robust exploitation for the class of deep q-network algorithms, it is easier to infer performance against DQN \cite{Mnih2013PlayingAW} and DDQN \cite{10.5555/3016100.3016191} if we compare to D3QN for obstacle avoidance. This is due to Xie et al. visualizing and comparing the performance gains against these algorithms in \cite{xie2017towards}. 

 The early exploration choices of guidance exploration showed significant improvements in future episodes, however improvements to the dynamic network's structure and prediction capabilities might be one area that could be explored further, as also suggested by \cite{Gou2019DQNWM}. Additionally, the sub population distribution of previous states in the priority replay memory may not be as appropriately estimated by a Gaussian mixture model as expected. An alternative distribution might be a more appropriate model of these previously visited states.

\section{Conclusion}
\label{sec:conclusion}

Application of artificial intelligence on a UAV system is a challenging process for obstacle avoidance, object detection, and self-learning have to be implemented with fine adjustments. Especially, reinforcement learning algorithm faces many limitations in 3-dimensional environment due to increasing number of states. We presented an improvement on the exploration algorithms to enhance the performance of the reinforcement learning framework in large state-space in multiple 3-D simulation environments. As an immediate application, vision based navigation of a UAV in a complex environment containing a moving object, was demonstrated with a higher performance compared to the existing algorithms. In such cases, where the locations and orientations of the UAV can collectively create an uncountable pixel based state-space, the traditional exploration algorithms may fail in completing the task.
The convergence exploration and guidance exploration methods were evaluated based on average rewards and number of steps taken per episode. Overall, our simulation results demonstrated that guidance exploration outperformed $\epsilon$-greedy exploration and convergence exploration. In the complex environment, the guidance exploration leads up to $200\%$ improvement in terms of average reward, and demonstrated significantly faster convergence in comparison to the state-of-the-art exploration algorithms. 
In similar environments, the proposed framework can result in a policy which allows for substantially larger amount of steps in contrast with the conventional obstacle avoidance algorithms that lead to collision in almost every episode.

Experiments of different exploration approaches results in faster learning, however, there is still room for improvement to stabilize the performance in case of exploration in more complex and dynamic environments. Additionally, exploring other real time object detection and tracking approaches such as incorporating additional and different non-static objects, different architectures and reward function changes will be investigated in the future. Moreover, we plan to apply our algorithms to a real UAV trained with a smaller number of episodes in order to gauge how significant improved exploration is to yielding more robust exploitation.

%
%


\begin{thebibliography}{22}
\providecommand{\natexlab}[1]{#1}
\providecommand{\url}[1]{\texttt{#1}}
\expandafter\ifx\csname urlstyle\endcsname\relax
  \providecommand{\doi}[1]{doi: #1}\else
  \providecommand{\doi}{doi: \begingroup \urlstyle{rm}\Url}\fi

\bibitem[Xie et~al.(2017)Xie, Wang, Markham, and Trigoni]{xie2017towards}
L.~Xie, S.~Wang, A.~Markham, and N.~Trigoni.
\newblock Towards monocular vision based obstacle avoidance through deep
  reinforcement learning.
\newblock In \emph{RSS 2017 workshop on New Frontiers for Deep Learning in
  Robotics}, 2017.

\bibitem[Lee et~al.(2016)Lee, Jung, and Shim]{inproceedings}
H.~Lee, S.~Jung, and D.~Shim.
\newblock Vision-based uav landing on the moving vehicle.
\newblock pages 1--7, 06 2016.
\newblock \doi{10.1109/ICUAS.2016.7502574}.

\bibitem[Michels et~al.(2005)Michels, Saxena, and Ng]{10.1145/1102351.1102426}
J.~Michels, A.~Saxena, and A.~Y. Ng.
\newblock High speed obstacle avoidance using monocular vision and
  reinforcement learning.
\newblock In \emph{Proceedings of the 22nd International Conference on Machine
  Learning}, ICML ’05, page 593–600, New York, NY, USA, 2005. Association
  for Computing Machinery.
\newblock ISBN 1595931805.
\newblock \doi{10.1145/1102351.1102426}.
\newblock URL \url{https://doi.org/10.1145/1102351.1102426}.

\bibitem[Preiss et~al.(2017)Preiss, H{\"o}nig, Sukhatme, and
  Ayanian]{Preiss2017CrazyswarmAL}
J.~A. Preiss, W.~H{\"o}nig, G.~S. Sukhatme, and N.~Ayanian.
\newblock Crazyswarm: A large nano-quadcopter swarm.
\newblock \emph{2017 IEEE International Conference on Robotics and Automation
  (ICRA)}, pages 3299--3304, 2017.

\bibitem[Asli et~al.(2019)Asli, Roghair, and Jannesari]{asli2019energy}
A.~Asli, J.~Roghair, and A.~Jannesari.
\newblock Energy-aware goal selection and path planning of uav systems via
  reinforcement learning.
\newblock \emph{arXiv preprint arXiv:1909.12217}, 2019.

\bibitem[Ma et~al.(2018)Ma, Wang, Niu, Wang, and Shen]{Ma2018ASR}
Z.~Ma, C.~Wang, Y.~Niu, X.~Wang, and L.~Shen.
\newblock A saliency-based reinforcement learning approach for a uav to avoid
  flying obstacles.
\newblock \emph{Robotics Auton. Syst.}, 100:\penalty0 108--118, 2018.

\bibitem[Smolyanskiy et~al.(2017)Smolyanskiy, Kamenev, Smith, and
  Birchfield]{Smolyanskiy2017TowardLA}
N.~Smolyanskiy, A.~Kamenev, J.~Smith, and S.~T. Birchfield.
\newblock Toward low-flying autonomous mav trail navigation using deep neural
  networks for environmental awareness.
\newblock \emph{2017 IEEE/RSJ International Conference on Intelligent Robots
  and Systems (IROS)}, pages 4241--4247, 2017.

\bibitem[Gou and Liu(2019)]{Gou2019DQNWM}
S.~Z. Gou and Y.~Liu.
\newblock Dqn with model-based exploration: efficient learning on environments
  with sparse rewards.
\newblock \emph{ArXiv}, abs/1903.09295, 2019.

\bibitem[Mnih et~al.(2013)Mnih, Kavukcuoglu, Silver, Graves, Antonoglou,
  Wierstra, and Riedmiller]{Mnih2013PlayingAW}
V.~Mnih, K.~Kavukcuoglu, D.~Silver, A.~Graves, I.~Antonoglou, D.~Wierstra, and
  M.~A. Riedmiller.
\newblock Playing atari with deep reinforcement learning.
\newblock \emph{ArXiv}, abs/1312.5602, 2013.

\bibitem[Wang et~al.(2015)Wang, Schaul, Hessel, Van~Hasselt, Lanctot, and
  De~Freitas]{wang2015dueling}
Z.~Wang, T.~Schaul, M.~Hessel, H.~Van~Hasselt, M.~Lanctot, and N.~De~Freitas.
\newblock Dueling network architectures for deep reinforcement learning.
\newblock \emph{arXiv preprint arXiv:1511.06581}, 2015.

\bibitem[Kahn et~al.(2017)Kahn, Villaflor, Pong, Abbeel, and
  Levine]{Kahn2017UncertaintyAwareRL}
G.~Kahn, A.~Villaflor, V.~Pong, P.~Abbeel, and S.~Levine.
\newblock Uncertainty-aware reinforcement learning for collision avoidance.
\newblock \emph{ArXiv}, abs/1702.01182, 2017.

\bibitem[Long et~al.(2017)Long, Fan, Liao, Liu, Zhang, and
  Pan]{Long2017TowardsOD}
P.~Long, T.~Fan, X.~Liao, W.~Liu, H.~Zhang, and J.~Pan.
\newblock Towards optimally decentralized multi-robot collision avoidance via
  deep reinforcement learning.
\newblock \emph{2018 IEEE International Conference on Robotics and Automation
  (ICRA)}, pages 6252--6259, 2017.

\bibitem[Wang et~al.(2017)Wang, Wang, Zhang, and Zhang]{Wang2017AutonomousNO}
C.~Wang, J.~Wang, X.~Zhang, and X.~Zhang.
\newblock Autonomous navigation of uav in large-scale unknown complex
  environment with deep reinforcement learning.
\newblock \emph{2017 IEEE Global Conference on Signal and Information
  Processing (GlobalSIP)}, pages 858--862, 2017.

\bibitem[Sutton and Barto(2018)]{Sutton1998}
R.~S. Sutton and A.~G. Barto.
\newblock \emph{Reinforcement Learning: An Introduction}.
\newblock The MIT Press, second edition, 2018.
\newblock URL \url{http://incompleteideas.net/book/the-book-2nd.html}.

\bibitem[Hasselt et~al.(2016)Hasselt, Guez, and
  Silver]{10.5555/3016100.3016191}
H.~v. Hasselt, A.~Guez, and D.~Silver.
\newblock Deep reinforcement learning with double q-learning.
\newblock In \emph{Proceedings of the Thirtieth AAAI Conference on Artificial
  Intelligence}, AAAI’16, page 2094–2100. AAAI Press, 2016.

\bibitem[Schaul et~al.(2015)Schaul, Quan, Antonoglou, and
  Silver]{Schaul2015PrioritizedER}
T.~Schaul, J.~Quan, I.~Antonoglou, and D.~Silver.
\newblock Prioritized experience replay.
\newblock \emph{CoRR}, abs/1511.05952, 2015.

\bibitem[Oh et~al.(2015)Oh, Guo, Lee, Lewis, and
  Singh]{Oh2015ActionConditionalVP}
J.~Oh, X.~Guo, H.~Lee, R.~L. Lewis, and S.~P. Singh.
\newblock Action-conditional video prediction using deep networks in atari
  games.
\newblock In \emph{NIPS}, 2015.

\bibitem[Shah et~al.(2017)Shah, Dey, Lovett, and Kapoor]{Shah2017AirSimHV}
S.~Shah, D.~Dey, C.~Lovett, and A.~Kapoor.
\newblock Airsim: High-fidelity visual and physical simulation for autonomous
  vehicles.
\newblock \emph{ArXiv}, abs/1705.05065, 2017.

\bibitem[Masadeh and Kamal(2018)]{masadeh18}
Z.~Masadeh, Ala'Eddin;~Wang and A.~E. Kamal.
\newblock Convergence-based exploration algorithm for reinforcement learning.
\newblock Electrical and Computer Engineering Technical Reports and White
  Papers~1, Iowa State University, Ames, IA, 2018.

\bibitem[Pathak et~al.(2017)Pathak, Agrawal, Efros, and
  Darrell]{pathakICMl17curiosity}
D.~Pathak, P.~Agrawal, A.~A. Efros, and T.~Darrell.
\newblock Curiosity-driven exploration by self-supervised prediction.
\newblock In \emph{ICML}, 2017.

\bibitem[Dadi et~al.(2013)Dadi, Venkatesh, Poornesh, Rao~L, and
  Kumar]{dadi13IJSCE}
H.~Dadi, P.~Venkatesh, P.~Poornesh, N.~Rao~L, and N.~Kumar.
\newblock Tracking multiple moving objects using gaussian mixture model.
\newblock \emph{International Journal of Soft Computing and Engineering
  (IJSCE)}, 3:\penalty0 114--119, 01 2013.

\bibitem[Chavan and Gengaje(2017)]{Chavan2017MultipleOD}
R.~Chavan and S.~R. Gengaje.
\newblock Multiple object detection using gmm technique and tracking using
  kalman filter.
\newblock 2017.

\end{thebibliography}


\begin{thebibliography}{6}
%
\bibitem{xie2017towards}
Xie, L., Wang, S., Markham, A. and Trigoni, N.: Towards monocular vision based obstacle avoidance through deep reinforcement learning. arXiv preprint arXiv:1706.09829 (2017).

\bibitem{inproceedings}
Lee, H., Jung, S. and Shim, D.H.: Vision-based UAV landing on the moving vehicle. In 2016 International conference on unmanned aircraft systems (ICUAS) (pp. 1-7). IEEE (2016).

\bibitem{10.1145/1102351.1102426}
Michels, J., Saxena, A. and Ng, A.Y.: High speed obstacle avoidance using monocular vision and reinforcement learning. In Proceedings of the 22nd international conference on Machine learning (pp. 593-600) (2005).

\bibitem{Preiss2017CrazyswarmAL}
Preiss, J.A., Honig, W., Sukhatme, G.S. and Ayanian, N. Crazyswarm: A large nano-quadcopter swarm. In 2017 IEEE International Conference on Robotics and Automation (ICRA) (pp. 3299-3304). IEEE (2017).

\bibitem{asli2019energy}
Asli, A.E., Roghair, J. and Jannesari, A.: Energy-aware Goal Selection and Path Planning of UAV Systems via Reinforcement Learning. arXiv preprint arXiv:1909.12217 (2019).

\bibitem{Ma2018ASR}
Ma, Z., Wang, C., Niu, Y., Wang, X. and Shen, L.: A saliency-based reinforcement learning approach for a UAV to avoid flying obstacles. Robotics and Autonomous Systems, 100, pp.108-118 (2018).

\bibitem{Smolyanskiy2017TowardLA}
Smolyanskiy, N., Kamenev, A., Smith, J. and Birchfield, S.: Toward low-flying autonomous MAV trail navigation using deep neural networks for environmental awareness. IEEE/RSJ International Conference on Intelligent Robots and Systems (IROS) (pp. 4241-4247). IEEE (2017).

\bibitem{Gou2019DQNWM}
Gou, S.Z. and Liu, Y.: DQN with model-based exploration: efficient learning on environments with sparse rewards. arXiv preprint arXiv:1903.09295 (2019).

\bibitem{Mnih2013PlayingAW}
Mnih, V., Kavukcuoglu, K., Silver, D., Graves, A., Antonoglou, I., Wierstra, D. and Riedmiller, M.: Playing atari with deep reinforcement learning. arXiv preprint arXiv:1312.5602 (2013).

\bibitem{kahn2018self}
Kahn, G., Villaflor, A., Ding, B., Abbeel, P. and Levine, S.: Self-supervised deep reinforcement learning with generalized computation graphs for robot navigation. In 2018 IEEE International Conference on Robotics and Automation (ICRA) (pp. 1-8). IEEE (2018).

\bibitem{halim2016artificial}
Halim, Z., Kalsoom, R., Bashir, S. and Abbas, G.: Artificial intelligence techniques for driving safety and vehicle crash prediction. Artificial Intelligence Review, 46(3), pp.351-387 (2016).

\bibitem{zhang2015towards}
Zhang, F., Leitner, J., Milford, M., Upcroft, B. and Corke, P.: Towards vision-based deep reinforcement learning for robotic motion control. arXiv preprint arXiv:1511.03791 (2015).

\bibitem{Sutton1998}
Sutton, R.S. and Barto, A.G.: Reinforcement learning: An introduction. MIT press (2018)

\bibitem{10.5555/3016100.3016191}
Van Hasselt, H., Guez, A. and Silver, D.: Deep reinforcement learning with double q-learning. In Proceedings of the AAAI conference on artificial intelligence Vol. 30, No. 1 (2016)

\bibitem{wang2015dueling}
Wang, Z., Schaul, T., Hessel, M., Hasselt, H., Lanctot, M. and Freitas, N: Dueling network architectures for deep reinforcement learning. In International conference on machine learning (pp. 1995-2003) (2016)

\bibitem{Girshick2013RCNN}
Girshick, R., Donahue, J., Darrell, T. and Malik, J.: Rich feature hierarchies for accurate object detection and semantic segmentation. In Proceedings of the IEEE conference on computer vision and pattern recognition pp. 580-587 (2014).

\bibitem{Liu2016SSDSS}
Liu, W., Anguelov, D., Erhan, D., Szegedy, C., Reed, S., Fu, C.Y. and Berg, A.C. Ssd: Single shot multibox detector. In European conference on computer vision (pp. 21-37). Springer, Cham (2016).

\bibitem{Redmon2016YouOL}
Redmon, J., Divvala, S., Girshick, R. and Farhadi, A.:You only look once: Unified, real-time object detection. In Proceedings of the IEEE conference on computer vision and pattern recognition pp. 779-788 (2016).

\bibitem{huang2017}
Huang, J., Rathod, V., Sun, C., Zhu, M., Korattikara, A., Fathi, A., Fischer, I., Wojna, Z., Song, Y., Guadarrama, S. and Murphy, K.: Speed/accuracy trade-offs for modern convolutional object detectors. In Proceedings of the IEEE conference on computer vision and pattern recognition pp. 7310-7311 (2017).


\bibitem{Lin2014MicrosoftCC}
Lin, T.Y., Maire, M., Belongie, S., Hays, J., Perona, P., Ramanan, D., Dollár, P. and Zitnick, C.L.: Microsoft coco: Common objects in context. In European conference on computer vision pp. 740-755. Springer, Cham (2017).

\bibitem{Howard2017MobileNetsEC}
Howard, A.G., Zhu, M., Chen, B., Kalenichenko, D., Wang, W., Weyand, T., Andreetto, M. and Adam, H.: Mobilenets: Efficient convolutional neural networks for mobile vision applications. arXiv preprint arXiv:1704.04861 (2017).


\bibitem{Kahn2017UncertaintyAwareRL}
Kahn, G., Villaflor, A., Pong, V., Abbeel, P. and Levine, S.: Uncertainty-aware reinforcement learning for collision avoidance. arXiv preprint arXiv:1702.01182 (2017).

\bibitem{Long2017TowardsOD}
Long, P., Fanl, T., Liao, X., Liu, W., Zhang, H. and Pan, J.: Towards optimally decentralized multi-robot collision avoidance via deep reinforcement learning.Iternational Conference on Robotics and Automation (ICRA) pp. 6252-6259  IEEE (2018).

\bibitem{Wang2017AutonomousNO}
Wang, C., Wang, J., Zhang, X. and Zhang, X.: Autonomous navigation of UAV in large-scale unknown complex environment with deep reinforcement learning. In 2017 IEEE Global Conference on Signal and Information Processing (GlobalSIP) pp. 858-862. IEEE (2017).

\bibitem {Schaul2015PrioritizedER}
Schaul, T., Quan, J., Antonoglou, I. and Silver, D.: Prioritized experience replay. arXiv preprint arXiv:1511.05952 (2015).

\bibitem{Oh2015ActionConditionalVP}
Oh, J., Guo, X., Lee, H., Lewis, R.L. and Singh, S.: Action-conditional video prediction using deep networks in atari games. Advances in neural information processing systems, 28, pp.2863-2871 (2015).

\bibitem{Shah2017AirSimHV}
Shah, S., Dey, D., Lovett, C. and Kapoor, A.: Airsim: High-fidelity visual and physical simulation for autonomous vehicles. In Field and service robotics pp. 621-635. Springer, Cham (2018).


\bibitem{masadeh18}
Masadeh, A.E., Wang, Z. and Kamal, A.E.: Convergence-based exploration algorithm for reinforcement learning (2018).

\bibitem{pathakICMl17curiosity}
Pathak, D., Agrawal, P., Efros, A.A. and Darrell, T.: Curiosity-driven exploration by self-supervised prediction. In Proceedings of the IEEE Conference on Computer Vision and Pattern Recognition Workshops pp. 16-17 (2017).

\bibitem{dadi13IJSCE}
Santosh, D.H.H., Venkatesh, P., Poornesh, P., Rao, L.N. and Kumar, N.A.: Tracking multiple moving objects using gaussian mixture model. International Journal of Soft Computing and Engineering (IJSCE), 3(2), pp.114-119 (2017).

\bibitem{Chavan2017MultipleOD}
Chavan, R. and Gengaje, S.R.: Multiple object detection using GMM technique and tracking using Kalman filter. Int. J. Comput. Appl, 172(3), pp.20-25 (2017).


\end{thebibliography}
\end{document}